\documentclass[lettersize,journal]{IEEEtran}
\usepackage{amsmath,amsfonts}
\usepackage{algorithmic}
\usepackage{array}
\usepackage[caption=false]{subfig}
\usepackage{cite}
\usepackage{textcomp}
\usepackage{stfloats}
\usepackage{url}
\usepackage{verbatim}
\usepackage{graphicx}
\usepackage{xcolor}
\usepackage{soul}
\usepackage{multirow}
\usepackage{orcidlink}
\usepackage{pifont}
\newcommand{\xmark}{\ding{55}}

\usepackage{balance}

\begin{document}
\title{MobilePlantViT: A Mobile-friendly Hybrid ViT for Generalized Plant Disease Image Classification}
\author{Moshiur Rahman Tonmoy\orcidlink{0009-0004-3628-8242}, Md. Mithun Hossain\orcidlink{0009-0001-4883-1802}, Nilanjan Dey\orcidlink{0000-0001-8437-498X}, ~\IEEEmembership{Senior Member,~IEEE}, M. F. Mridha\orcidlink{0000-0001-5738-1631}, ~\IEEEmembership{Senior Member,~IEEE}
\thanks{Moshiur Rahman Tonmoy is with the Department of Computer Science and Engineering, University of Asia Pacific, Dhaka 1205, Bangladesh (email: moshiurtonmoy.bb@gmail.com)}
\thanks{Md. Mithun Hossain is with the Department of Computer Science and Engineering, Bangladesh University of Business and Technology, Dhaka 1216, Bangladesh (email: mhosen751@gmail.com)}
\thanks{Nilanjan Dey is with the Department of Computer Science and Engineering, Techno International New Town, Kolkata 700156, India (email: nilanjan.dey@tint.edu.in)}
\thanks{M. F. Mridha is with the Department of Computer Science, American International University-Bangladesh, Dhaka 1229, Bangladesh (email: firoz.mridha@aiub.edu)}
}



\maketitle

\begin{abstract}
Plant diseases significantly threaten global food security by reducing crop yields and undermining agricultural sustainability. AI-driven automated classification has emerged as a promising solution, with deep learning models demonstrating impressive performance in plant disease identification. However, deploying these models on mobile and edge devices remains challenging due to high computational demands and resource constraints, highlighting the need for lightweight, accurate solutions for accessible smart agriculture systems. To address this, we propose MobilePlantViT, a novel hybrid Vision Transformer (ViT) architecture designed for generalized plant disease classification, which optimizes resource efficiency while maintaining high performance. Extensive experiments across diverse plant disease datasets of varying scales show our model's effectiveness and strong generalizability, achieving test accuracies ranging from 80\% to over 99\%. Notably, with only 0.69 million parameters, our architecture outperforms the smallest versions of MobileViTv1 and MobileViTv2, despite their higher parameter counts. These results underscore the potential of our approach for real-world, AI-powered automated plant disease classification in sustainable and resource-efficient smart agriculture systems. All codes will be available in the GitHub repository: \url{https://github.com/moshiurtonmoy/MobilePlantViT}
\end{abstract}


\section{Introduction}
\IEEEPARstart{P}{lant} diseases represent a significant threat to global food security, impacting crop output and endangering the livelihoods of millions of people. According to the Food and Agriculture Organization (FAO), up to 40\% of global crop production is lost annually due to pests and diseases, amounting to billions of dollars in economic losses and exacerbating food scarcity challenges \cite{FAO2018}. Automated and accurate identification of these diseases is essential to mitigate crop damage and ensure sustainable agricultural productivity, and deep learning (DL) has revolutionized the field of automated plant disease recognition \cite{Simonyan2014, szegedy2015going} over the past decade. Convolutional neural network (CNN)--based models have demonstrated notable success in extracting discriminative features from complex leaf images, achieving competitive performance in classification tasks \cite{mohanty2016using, sladojevic2016deep}. In recent years, Vision Transformers (ViTs) have emerged as a compelling alternative to CNNs \cite{khan2022transformers} and have shown remarkable performance gains across a wide range of computer vision tasks, including classification, suggesting their potential applicability in plant disease diagnosis \cite{chen2021empirical}.

However, deploying high-performance DL models to resource-constrained devices remains a major challenge. For instance, the high computational and memory demands of ViTs can be prohibitive in low-power scenarios \cite{liu2024edvit} despite their state-of-the-art performance. Likewise, conventional CNN backbones also face trade-offs between model capacity and hardware limitations, spurring researchers into lightweight models. There has been significant progress toward developing compact models including ViTs \cite{khan2022transformers}, and models such as MobileViTv1\cite{mehta2022mobilevit} and MobileViTv2\cite{mehta2023separable} are the prime examples. They incorporate efficient mechanisms for compressed feature representations and streamlined designs to narrow the performance gap between standard ViT and CNN counterparts \cite{cao2023ghostvit, chen2022mobileformer}. Nonetheless, these methods can remain computationally intensive in ultra-low-power devices, underscoring the need for optimizations that balance accuracy and maximum efficiency in ensuring precision agriculture accessibility globally.

In the context of scaling AI-based solutions for global precision agriculture advancement across underrepresented farming communities, tailored lightweight architectures become even more critical for cost-efficient automated agricultural tasks \cite{SHARMA2024100292}. As high-end smart agriculture tools and automated systems often put significant financial barriers to users from low-income regions, introducing on-device smart systems integrated with tailored lightweight AI models demanding minimum computational overhead could be a feasible choice. With this motivation, we investigate lightweight yet generalized DL models in the plant disease classification domain and propose MobilePlantViT, a hybrid ViT architecture tailored for accurate plant disease image classification maintaining well balance between performance and efficiency. The key contributions of this study can be summarized as follows:

\begin{itemize}
    \item We propose a generalized lightweight ViT for accurate plant disease image classification, with only 0.69 million parameters
    \item Our model achieved competitive accuracy across extensive experiments on multiple plant disease datasets
    \item Despite having fewer parameters, our model outperformed equivalent lightweight ViTs with higher parameter counts
\end{itemize}
The rest of this article is organized as follows: Section \ref{sec:lr}
summarizes past works, Section \ref{sec:method} presents methodology, Section \ref{sec:results} provides the experimental results and discusses the key findings. Finally, Section \ref{sec:conclusion} concludes this study with future directions.

\section{Related Works}\label{sec:lr}
Over the years, researchers have focused on various aspects of automated plant disease recognition systems, particularly with the effectiveness of CNN-based architectures. Rehman et al. \cite{rehman2024leveraging} employed CNNs to identify diseases in vegetables while overcoming obstacles like limited datasets, though limiting its scalability for underrepresented crops. Similarly, Demilie \cite{demilie2024plant} compared multiple CNNs and reported up to 99.60\% accuracy via ensemble approaches, however, ensemble methods increase computational overhead, potentially constraining edge deployment. CNNs were further shown to be effective by Gupta et al. \cite{singla2023plant} and Bezabih et al. \cite{bezabh2023cpdccnn}, who combined VGG16 and AlexNet to classify pepper diseases with nearly perfect accuracy, but these relatively older architectures can be computationally heavier compared to modern variants. Additionally, Hassan and Maji \cite{hassan2022plant}, and Mzoughi and Yahiaoui \cite{mzoughi2023deep} presented sophisticated CNN models that included segmentation methods to enhance accuracy while lowering complexity, yet such segmentation-based approaches often add extra preprocessing steps. To further improve classification performance, Wang et al. \cite{wang2021tcnn} and Nagaraju and Chawla \cite{nagaraju2021plant} proposed T-CNN and DCNN-19, respectively, although these architectures still risk high inference times on low-power devices. According to Ojo and Zahid \cite{ojo2023improving}, class imbalance and data inconsistencies can be effectively reduced by combining ResNet-50 with preprocessing methods like CLAHE and GAN-based resampling, but these additional steps may complicate the pipeline and increase total computational overhead. Furthermore, Chen et al. \cite{chen2020identifying} illustrated the advantages of deep transfer learning using a modified MobileNet-V2 to improve the detection of subtle lesion symptoms. The DIC-Transformer presented by Zeng et al. \cite{zeng2024dic} achieves 85.4\% accuracy while delivering comprehensive disease descriptions by combining Faster R-CNN for disease identification with a Transformer for picture caption creation. Nonetheless, this heavy two-stage pipeline can be computationally intensive for cost-effective applications. To improve feature extraction and prediction accuracy over models such as FCN-8s and DeepLabv3, Kalpana et al. \cite{kalpana2024plant} suggested an ensemble of Swin transformers and residual convolutional networks, although ensemble approaches typically demand greater memory usage and training time. Zhu et al. \cite{zhu2023crop} introduced MSCVT, a hybrid model combining CNNs and ViT that achieved 99.86\% accuracy on PlantVillage with reduced parameter complexity. Teki et al. \cite{teki2023comparison} applied ViT and Shifted Window Transformer to thermal images of paddy leaves, achieving accuracies of 94\% and 98\%, respectively, but the requirement for thermal imaging devices and specialized hardware could restrict broad adoption. Borhani et al. \cite{borhani2022deep} investigated lightweight ViT-based techniques that, despite being slower than conventional CNNs, achieved higher accuracy, underscoring a persistent trade-off between performance and resource usage in Transformer-based approaches. MFF-ADNet, which combines ViT, VGG16, and a variational autoencoder, was introduced by Bathula et al. \cite{nagachandrika2024automatic} and outperforms current techniques by 7.18\%, though this integration of multiple components can inflate memory requirements and training complexity. AppViT, a lightweight hybrid model that merges convolutional blocks with multi-head self-attention, was proposed by Wasi Ullah et al. \cite{ullah2024efficient} and achieved 96.38\% precision on the Plant Pathology 2021-FGVC8 dataset. However, multi-head attention can be computationally heavy on extremely resource-limited devices. Chen et al. \cite{tabbakh2023deep} merged transfer learning with a vision transformer to attain validation accuracies up to 99.86\%, but it remains unclear how robust these models are under field conditions with non-uniform lighting or overlapping leaves. Li et al. \cite{zhang2024plant} introduced PDLM-TK, which combines tensor features with knowledge distillation and lightweight residual blocks and obtained 96.19\% accuracy and a 94.94\% f1 score. Thakur et al. \cite{thakur2023vision} combined ViTs with CNNs and produced 98.86\% accuracy on the PlantVillage dataset with just 0.85M parameters. Finally, Muthireddy et al. \cite{muthireddy2022plant} and Chougui et al. \cite{chougui2022plant} demonstrated that adding handcrafted features and attention mechanisms improves classifier performance. Nonetheless creating such handcrafted features can be labor-intensive and may not generalize well to complex or newly emerging plant diseases. 

Overall, past studies have highlighted the efficacy of CNN and ViT-based methods for various aspects of plant disease classification. However, these studies focused predominantly on maximizing accuracy and were often limited to specific plant species. Deploying separate DL models for different crops would be impractical and costly, underscoring the need for a tailored, lightweight architecture with strong generalization capabilities across multiple crops.

\section{Methodology}\label{sec:method}





\begin{figure*}
    \centering
    \includegraphics[width=18cm, trim=0.0cm 67.5cm 8.50cm 0.0cm, clip]{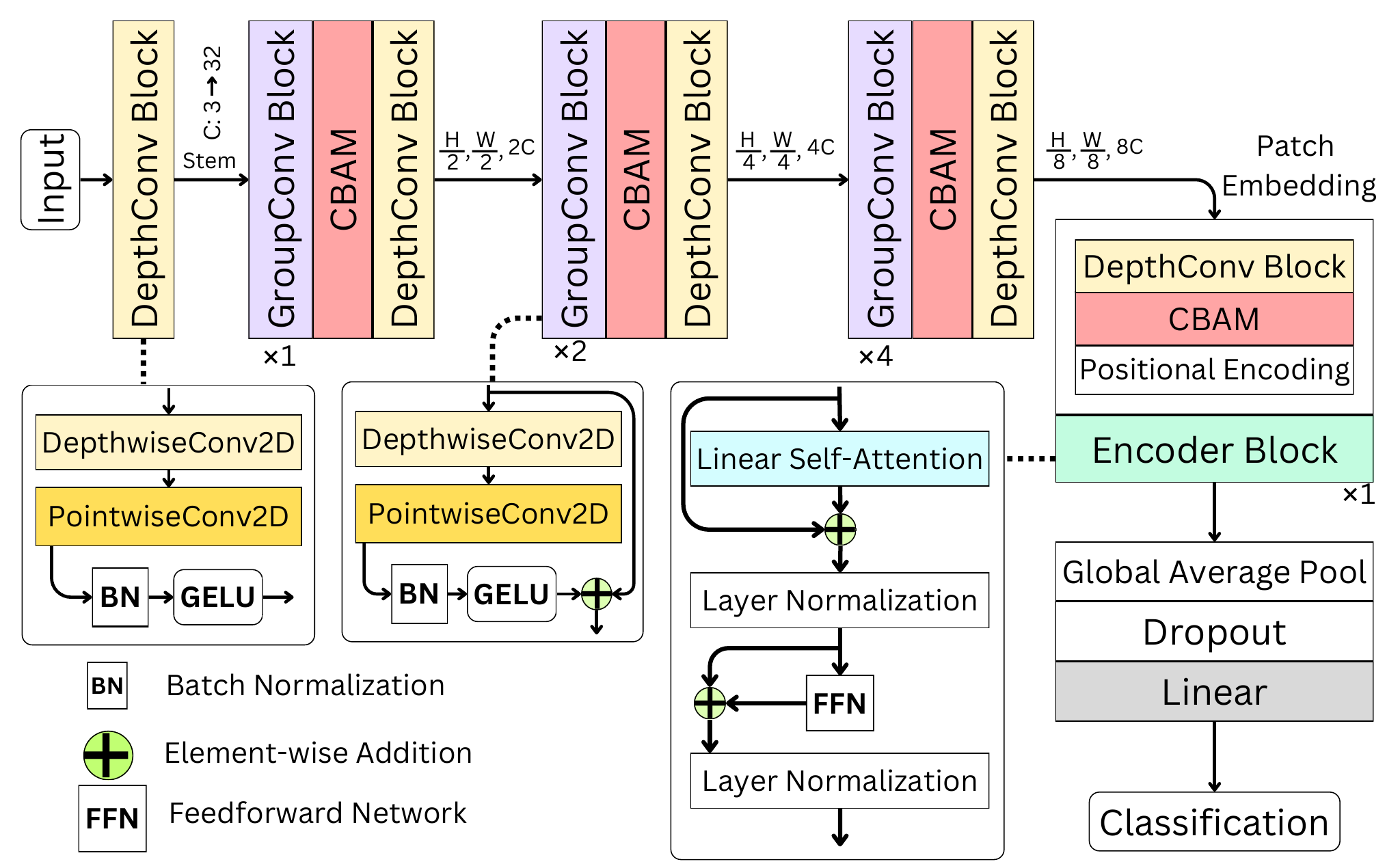}
    \caption{Outline of the proposed MobilePlantViT. It consists of a ×1-×2-×4-×1 combination of GroupConv and Encoder blocks. The first DepthConv block acts as the stem layer, expanding the initial channel dimension from 3 to 32. The last DepthConv block serves as the patch embedding layer, while all intermediate DepthConv blocks function as spatial pooling layers with dimension reduction and channel expansion}
    \label{fig:model-arch}
\end{figure*}

Our proposed architecture comprises multiple components as illustrated in Figure \ref{fig:model-arch}. It begins with a DepthConv block as the stem layer which increases the channel dimension from 3 to 32 along with the very first feature extraction operation. A DepthConv block essentially employs Depthwise Separable Convolution \cite{Chollet_2017_CVPR}, an efficient variant of the standard convolution which works in two steps: first, each input channel is convolved independently using a single spatial filter (Depthwise Convolution). Then, a $1\times1$ convolution (Pointwise Convolution) is applied to combine the outputs from the depthwise step and mix information across channels. Next, the GroupConv block comprises Group Convolution which can be considered as a generalized form of Depthwise Separable Convolution where each input channels are divided into user-defined groups, and each group is convolved separately using a dedicated set of filters, and the Depthwise Convolution is a special case where the number of groups equals the number of input channels. We set half the input channels as groups in each GroupConv block. Additionally, we imposed a residual connection within each block for effective information flow. These multi-step and residual operations significantly reduce the computational burden of standard convolution while enabling efficient feature extraction in deep networks.

The CBAM stands for Convolutional Block Attention Module \cite{Woo_2018_ECCV} which enhances feature representation by sequentially applying channel attention and spatial attention, allowing the network to focus on important information while suppressing less significant features. The channel attention module emphasizes identifying the important features by learning a channel-wise attention map by utilizing global average pooling (GAP) \& global max pooling (GMP) of the feature maps followed by a multilayer perceptron (MLP) with sigmoid activation to generate attention weights. Mathematically, let $F$ be the feature maps, then the channel attention $M_c(F)$ is:

\begin{equation}
    M_c(F) = \sigma(W_1(\delta(W_0(GAP(F))) + W_1(\delta(W_0(GMP(F))))))
\end{equation}
\begin{equation}
    F' = M_c(F) \odot F
\end{equation}
\noindent where $\sigma$ is sigmoid function $(\frac{1}{1 + e^{-x}})$, $\delta$ means ReLU activation which is $max(0, x)$, $W_1$ and $W_0$ are learnable MLP weights, $\odot$ is element-wise multiplication, and $F'$ is the refined feature maps with channel attention.
Similarly, the spatial attention module focuses on important regions in the feature map using a $7\times7$ convolution after applying max and average pooling across channels to generate a spatial attention map. Mathematically, let $F'$ be the refined feature maps after the channel attention module, then the spatial attention $M_s(F')$ is:

\begin{equation}
    M_s(F') = \sigma(f^{7\times7}([AvgPool(F'); MaxPool(F')]))
\end{equation}
\begin{equation}
    F'' = M_s(F') \odot F'
\end{equation}
\noindent where $f^{7\times7}$ is a $7\times7$ convolution, and $ F''$ is the refined feature maps output after CBAM. Then, we employed patch embedding with positional encoding over the extracted feature maps $(\frac{H}{8},\frac{W}{8}, 8C)$ for the encoder block. The patch embedding splits the input feature maps into fixed-size patches and transforms each patch into a high-dimensional vector, forming a sequence of image patches. We used a DepthConv block with stride and kernel size equal to patch size to split the features into patches. In addition, we integrated a CBAM block into the patch extraction process. Positional encoding is also added to the patch embeddings to retain spatial information, as encoders do not naturally capture spatial relationships. Then, instead of the classic Transformer encoder, we integrated a lightweight encoder block comprising a linear self-attention block instead of the multi-head self-attention. We adapted the linear self-attention mechanism from the MobileViTv2 \cite{mehta2023separable} as it enables efficient feature attention with linear complexity, suitable for resource-constraint devices. Given an input representation \( X \), the encoder block computes the attention output:

\begin{equation}
X' = \text{LN}(X + \text{LinearAttention}(X))
\end{equation}

\noindent where LN represents layer normalization. In LinearAttention block, given an input sequence \( X \in \mathbb{R}^{L \times d} \), where \( L \) is the sequence length, and \( d \) is the embedding dimension, we compute the query ($Q$), key ($K$), and value ($V$) representations as:

\begin{equation}
Q, K, V = W_{qkv} X + b_{qkv}
\end{equation}

\noindent where \( W_{qkv} \in \mathbb{R}^{(1+2d) \times d} \) is a learnable weight matrix, and \( b_{qkv} \) is a bias term. The \( Q \in \mathbb{R}^{L \times 1} \), and \( K, Q \in \mathbb{R}^{L \times d} \) are obtained via linear projection. The attention or context scores are computed using a softmax function applied along the sequence length, and then go through a dropout regularization layer:

\begin{equation}
\alpha = \text{{Softmax}(Q)}
\end{equation}
\begin{equation}
    \text{Softmax}(x_i) = \frac{e^{x_i}}{\sum_{j=1}^{n} e^{x_j}}
\end{equation}
\noindent where $x_i$ is the $i$-th input value and $n$ is the total number of input values. Thus, \( \alpha \in \mathbb{R}^{L \times 1} \) represents the normalized attention scores. Next, we compute the context vector as:

\begin{equation}
C = \sum_{i=1}^{L} \alpha_i \cdot K_i
\end{equation}

\noindent where \( C \in \mathbb{R}^{1 \times d} \) is the aggregated $K$ representation weighted by the attention scores. The output of the attention mechanism is then obtained by applying the GELU activation on the $V$, followed by element-wise multiplication with the expanded context vector. Next, a linear transformation is applied:

\begin{equation}
\hat{O} = W_{\text{out}} (\text{GELU}(V) \odot C) + b_{\text{out}}
\end{equation}

\noindent Here, GELU activation is $x \Phi(x)$ where the $\Phi(x)$ is the standard gaussian cumulative distribution function \cite{gelu}. \( W_{\text{out}} \in \mathbb{R}^{d \times d} \) and \( b_{\text{out}} \) are trainable parameters. The $X'$ is then processed by a feedforward network (FFN) of two linear layers with GELU and dropout regularization.

\begin{equation}
\text{FFN}(X') = \left(W_2 \cdot \text{GELU} (W_1 X' + b_1) + b_2 \right)
\end{equation}
\begin{equation}
X'' = \text{LN}(X' + \text{FFN}(X'))
\end{equation}

\noindent where \( W_1 \in \mathbb{R}^{d \times d_{ff}} \) and \( W_2 \in \mathbb{R}^{d_{ff} \times d} \) are the feedforward layer weights, and \( d_{ff} \) is the hidden dimension of the FFN. Residual connections are applied at both the attention and feedforward layers. The encoder gives us the output representation \( X'' \in \mathbb{R}^{L \times d} \). To obtain a global feature representation, we apply GAP along the $L$ (sequence length):

\begin{equation}
Z = \frac{1}{L} \sum_{i=1}^{L} X''_i
\end{equation}

\noindent where \( Z \in \mathbb{R}^{B \times d} \) represents the pooled feature vector and $B$ is the batch size. Dropout is applied to prevent overfitting. Lastly, a linear layer is applied to obtain the predicted logits for classification:

\begin{equation}
\hat{y} = W_{\text{c}} Z' + b_{\text{c}}
\end{equation}

\noindent where \( W_{\text{c}} \in \mathbb{R}^{d \times C} \) and \( b_{\text{c}} \in \mathbb{R}^{C} \) are the learnable parameters of the classifier and the output \( \hat{y} \in \mathbb{R}^{B \times C} \) represents the predicted logits which eventually go through a softmax to obtain class probabilities and we pick the class with the maximum value:

\begin{equation}
p_i = \text{Softmax}(\hat{y})
\end{equation}
\begin{equation}
\hat{c} = \arg\max_{i} p_i
\end{equation}

where \( p_i \) represents the probability of class \( i \) ensuring $\sum_{i=1}^{C} p_i = 1$, and \( \hat{c} \) is the predicted class.

\section{Results and Discussion}\label{sec:results}
In this section, experimental setup and datasets are presented. Subsequently, the experimental results on the performance of the model are discussed highlighting the key findings from various aspects. 

\subsection{Experimental Data and Setup}\label{subsec:data}
We have employed 4 datasets ranging from large-scale to small-scale: PlantVillage \cite{plantvill-latest}, CCMT \cite{ccmt-dataset}, Sugarcane \cite{sugarcane-dataset}, and Coconut \cite{coconut-dataset}. The PlantVillage dataset consists of 54303 healthy and unhealthy leaf images divided into 38 categories by species and diseases. The CCMT dataset contains 6549, 7508, 5389, and 5435 disease and healthy raw images of Cashew, Cassava, Maize, and Tomato. Each of them has 5 classes except for Maize which has 7 classes. The Sugarcane dataset includes 6748 high-resolution leaf images classified into 9 disease categories and the Coconut dataset comprises 5798 images across 5 disease categories. Datasets were split into training, validation, and test sets in a 70-15-15 ratio. We applied learning rate reduction and early stopping based on validation accuracy. If accuracy did not improve for 10 consecutive epochs, the learning rate was reduced by half, and if the non-improvement trend continued for 50 epochs, training was halted and the best weights were retrieved. We implemented our model using PyTorch and all experiments were conducted in Google Colab's notebook environment with an NVIDIA Tesla T4 GPU. Table \ref{tab:testbed} summarizes the experimental setup, including augmentation techniques applied to the training images. All data were normalized using a mean of (0.485, 0.456, 0.406) and a standard deviation of (0.229, 0.224, 0.225).

\begin{table}[htbp]
    \centering
    \caption{Experimental Setup and Data Augmentation Summary}
    \label{tab:testbed}
    \begin{tabular}{lc}
        \hline\hline
        \textbf{Attribute} & \textbf{Value}\\ \hline
        Input Shape & $ 3 \times 224 \times 224$ \\
        Batch Size & 64 \\
        Embed Dimension & 256\\
        FFN Dimension & 512\\
        Patch Size & 00\\
        Initial Learning Rate & $1e-3$ \\
        Learning Rate Reduction Rate & 50\% \\
        Minimum Learning Rate & $1e-7$\\
        Learning Rate Reduction Patience & 10\\
        Validation Accuracy Threshold & $1e-5$ \\ 
        Early Stopping Patience & 50\\
        Weight Decay & $1e-3$ \\
        Encoder Dropout Rate & 30\%\\
        Classifier Dropout Rate & 20\%\\
        Optimizer & Adam \\
        Loss & Categorical Crossentropy \\\hline
        Horizontal Flip &  50\%\\ 
         Random Rotate 90° & 50\%\\
         Shift, Scale, Rotate (limits: 0.05, 0.05, 30°) &  50\%  \\
         Random Gamma & 20\%\\
         Random Brightness/Contrast &  30\% \\ 
         RGB Shift (shift limits: 15) & 30\%  \\
         CLAHE (clip limit: 4.0) & 30\%\\
        \hline\hline
    \end{tabular}
\end{table}

\subsection{Performance Analysis}\label{subsec:perfor-analysis}

\begin{table*}[htpb]
    \centering
    \caption{Performance summary of the proposed model over multiple datasets, evaluated in macro and weighted average}
    \label{tab:performance-table}
    \begin{tabular}{lcccc|cc|cc|cc}
        \hline\hline
        \multirow{2}{*}{\textbf{Data}} 
        & \multirow{2}{*}{\textbf{Test Size}} & \multirow{2}{*}{\textbf{Accuracy (\%)}} 
        & \multicolumn{2}{c}{\textbf{Precision (\%)}} & \multicolumn{2}{c}{\textbf{Recall (\%)}} & \multicolumn{2}{c}{\textbf{F1-score}} & \multicolumn{2}{c}{\textbf{AUC (OvR)}} \\ 
        \cline{4-11}
        & & & \textbf{Macro} & \textbf{Weighted} & \textbf{Macro} & \textbf{Weighted} & \textbf{Macro} & \textbf{Weighted} & \textbf{Macro} & \textbf{Weighted} \\ 
        \hline
        {PlantVillage}  & 8179 & 99.57 & 99.47 & 99.58 & 99.40 & 99.57 & 0.9943 & 0.9957 & 0.9999 & 0.9999 \\
        {Coconut}  & 874 & 99.20 & 99.47 & 99.21 & 99.35 & 99.20 & 0.9941 & 0.9920 & 0.9999 & 0.9999 \\
        {Sugarcane}  & 1022 & 92.76 & 88.81 & 92.92 & 88.85 & 92.76 & 0.8870 & 0.9274 & 0.9976 & 0.9984 \\
        {CCMT-Cashew} & 987 & 95.04 & 95.71 & 95.02 & 95.78 & 95.04 & 0.9574 & 0.9502 & 0.9939 & 0.9927 \\
        {CCMT-Cassava} & 1131 & 94.34 & 94.46 & 94.40 & 94.41 & 94.36 & 0.9442 & 0.9436 & 0.9952 & 0.9944 \\  
        {CCMT-Maize}  & 802 & 81.05 & 82.87 & 81.41 & 81.42 & 81.05 & 0.8185 & 0.8110 & 0.9731 & 0.9668 \\
        {CCMT-Tomato} & 872 & 80.05 & 79.58 & 79.86 & 76.36 & 80.05 & 0.7754 & 0.7973 & 0.9517 & 0.9414 \\        
        \hline\hline
    \end{tabular}
\end{table*}

\begin{table*}[htpb]
    \centering
    \caption{Summary of the proposed model's performance with PlantVillage pre-trained weights}
    \label{tab:pretrained-performance-table}
    \begin{tabular}{lccc|cc|cc}
        \hline\hline
        \multirow{2}{*}{\textbf{Data}} 
        & \multirow{2}{*}{\textbf{Accuracy (\%)}} 
        & \multicolumn{2}{c}{\textbf{Precision (\%)}} & \multicolumn{2}{c}{\textbf{Recall (\%)}} & \multicolumn{2}{c}{\textbf{F1-score}} \\ 
        \cline{3-8}
        & & \textbf{Macro} & \textbf{Weighted} & \textbf{Macro} & \textbf{Weighted} & \textbf{Macro} & \textbf{Weighted} \\ 
        \hline
        {CCMT-Cashew} & 96.45 (1.48$\uparrow$) & 97.02 (1.37$\uparrow$) & 96.45 (1.51$\uparrow$) & 96.96 (1.23$\uparrow$) & 96.45 (1.48$\uparrow$) & 0.9698 (1.24\%$\uparrow$) & 0.9645 (1.51\%$\uparrow$) \\
        {CCMT-Cassava} & 95.23 (0.94$\uparrow$) & 95.48 (1.09$\uparrow$) & 95.25 (0.91$\uparrow$) & 95.63 (1.30$\uparrow$) & 95.23 (0.94$\uparrow$) & 0.9555 (1.20\%$\uparrow$) & 0.9524 (0.93\%$\uparrow$) \\
        {CCMT-Maize}  & 84.29 (4.00$\uparrow$) & 85.67 (3.38$\uparrow$) & 84.31 (3.56$\uparrow$) & 85.28 (4.74$\uparrow$) & 84.29 (4.00$\uparrow$) & 0.8541 (3.56\%$\uparrow$) & 0.8425 (3.88\%$\uparrow$) \\
        {CCMT-Tomato} & 83.60 (4.50$\uparrow$) & 82.90 (4.20$\uparrow$) & 83.40 (4.44$\uparrow$) & 80.98 (6.12$\uparrow$) & 83.60 (4.50$\uparrow$) & 0.8188 (4.34\%$\uparrow$) & 0.8345 (4.67\%$\uparrow$) \\
        \hline\hline
    \end{tabular}
\end{table*}

\begin{figure*} 
    \centering
  \subfloat[CCMT-Cashew\label{1a}]{%
       \includegraphics[width=0.24\linewidth]{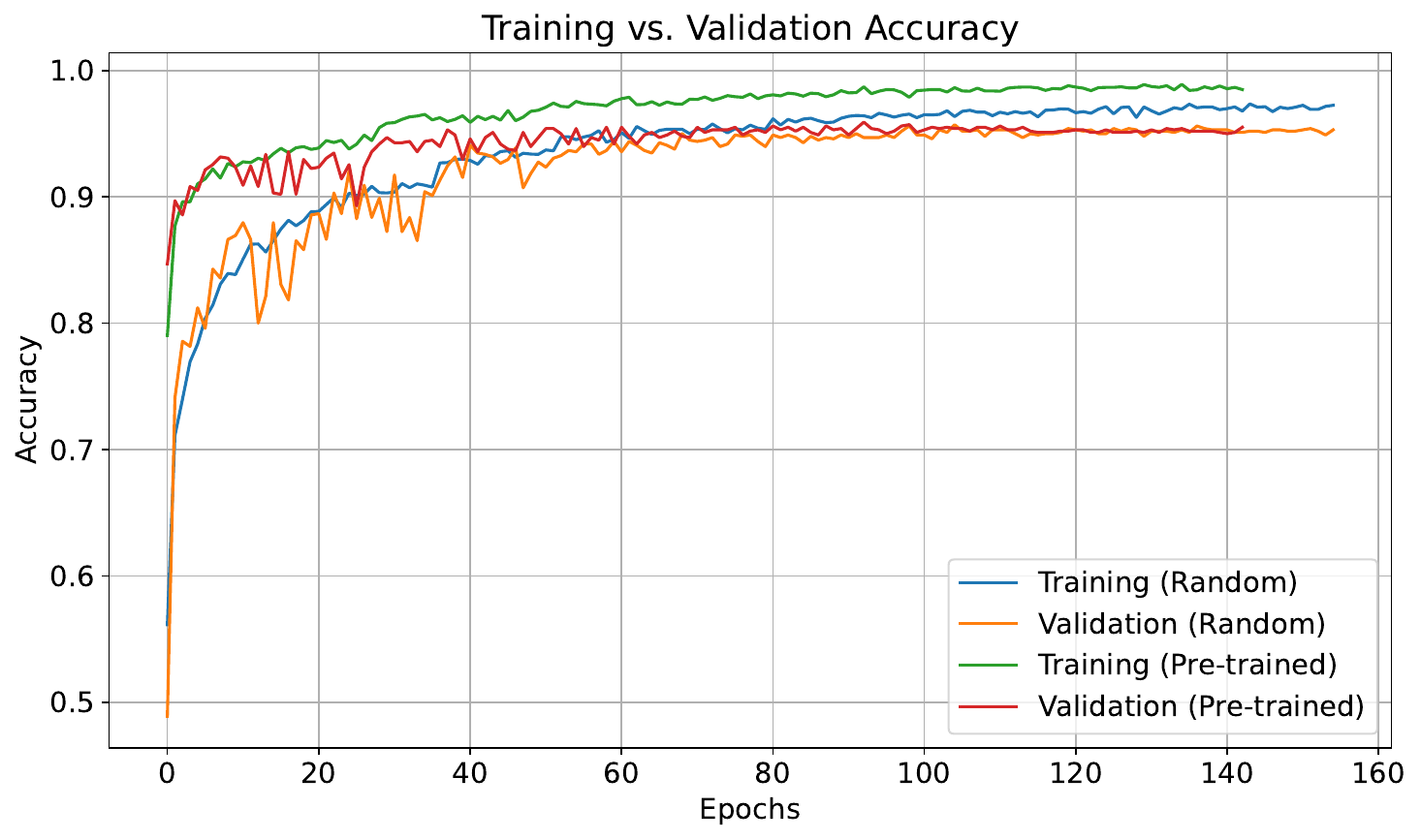}}
    \hfill
  \subfloat[CCMT-Cassava\label{1b}]{%
        \includegraphics[width=0.24\linewidth]{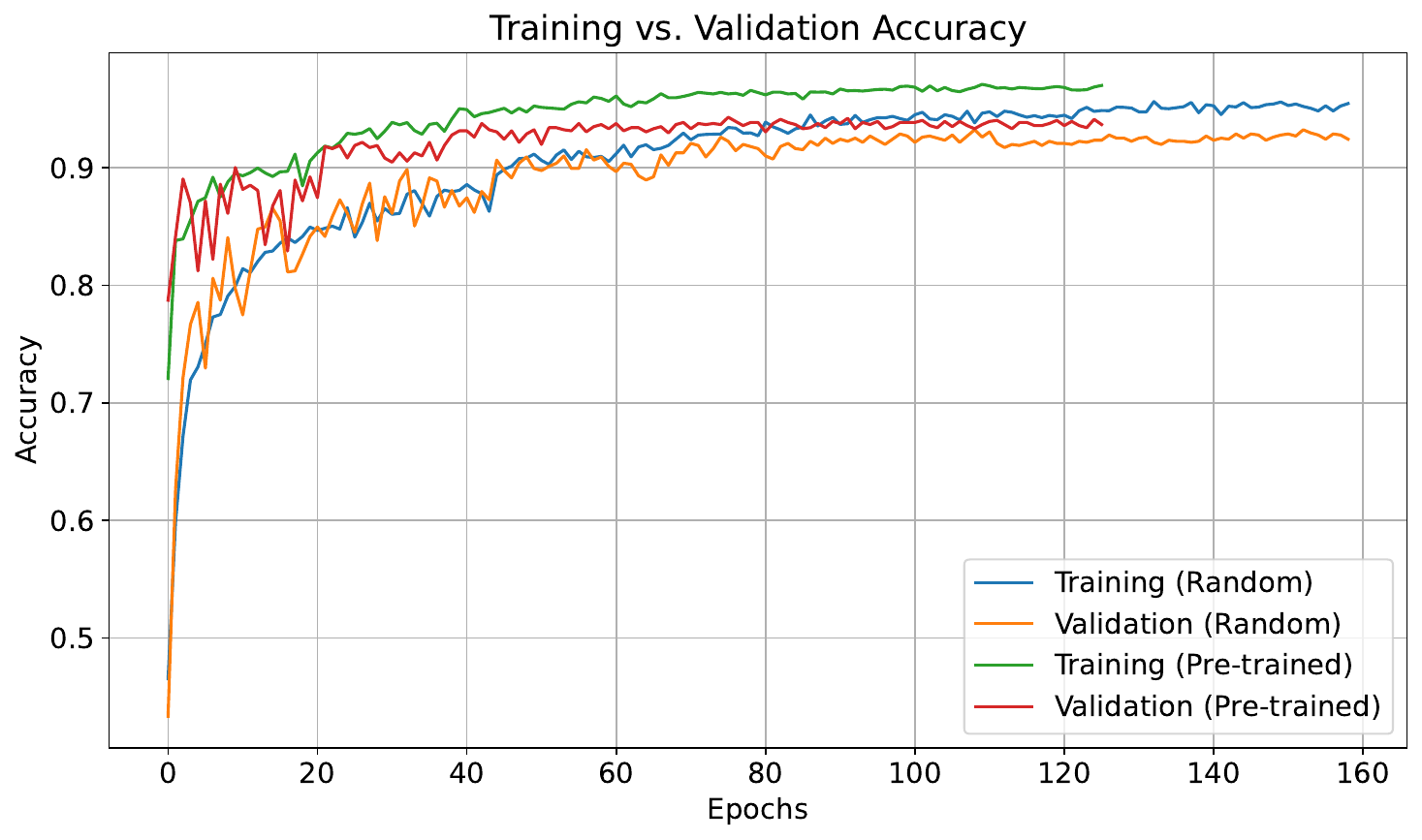}}
    \hfill
  \subfloat[CCMT-Maize\label{1c}]{%
        \includegraphics[width=0.24\linewidth]{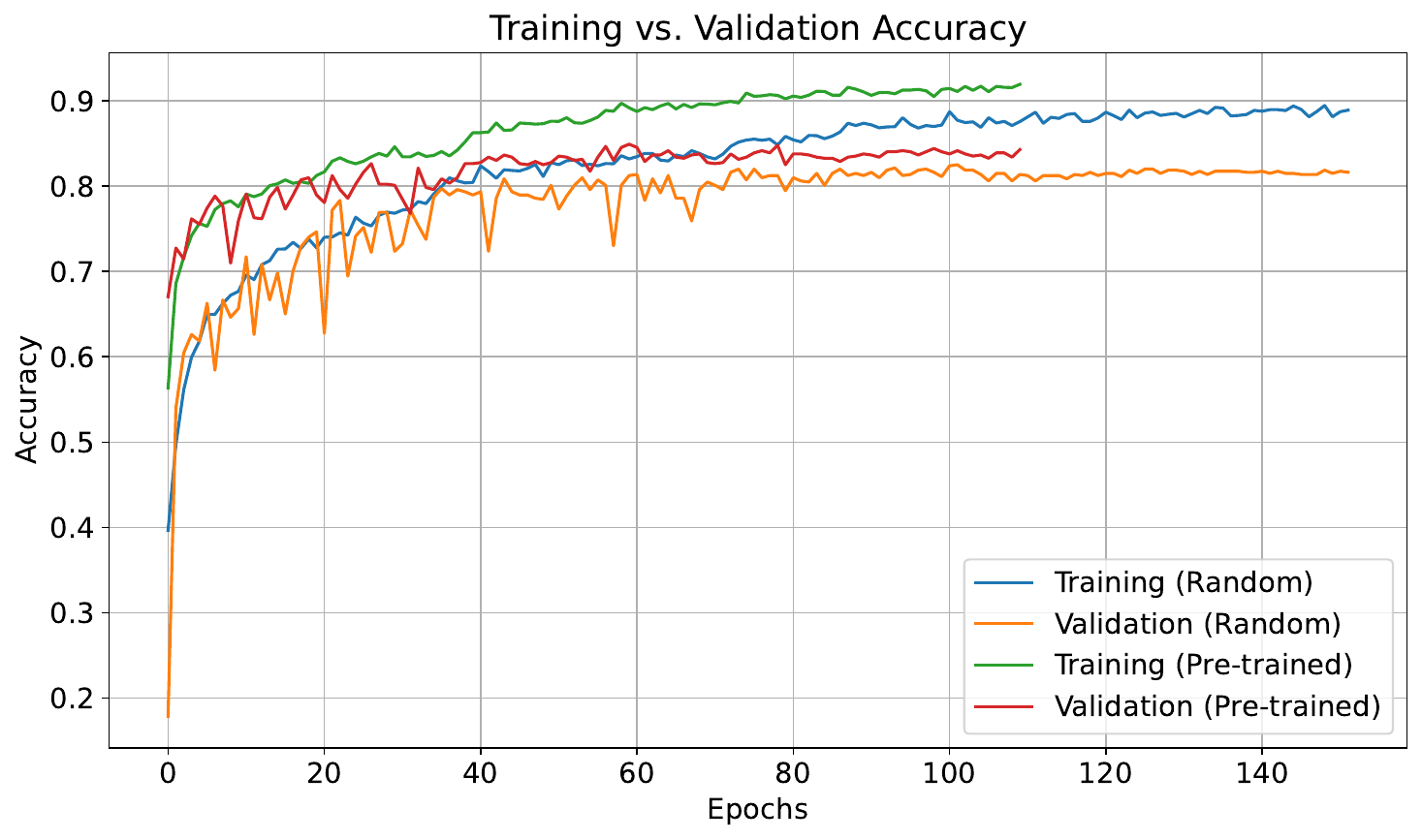}}
    \hfill
  \subfloat[CCMT-Tomato\label{1d}]{%
        \includegraphics[width=0.24\linewidth]{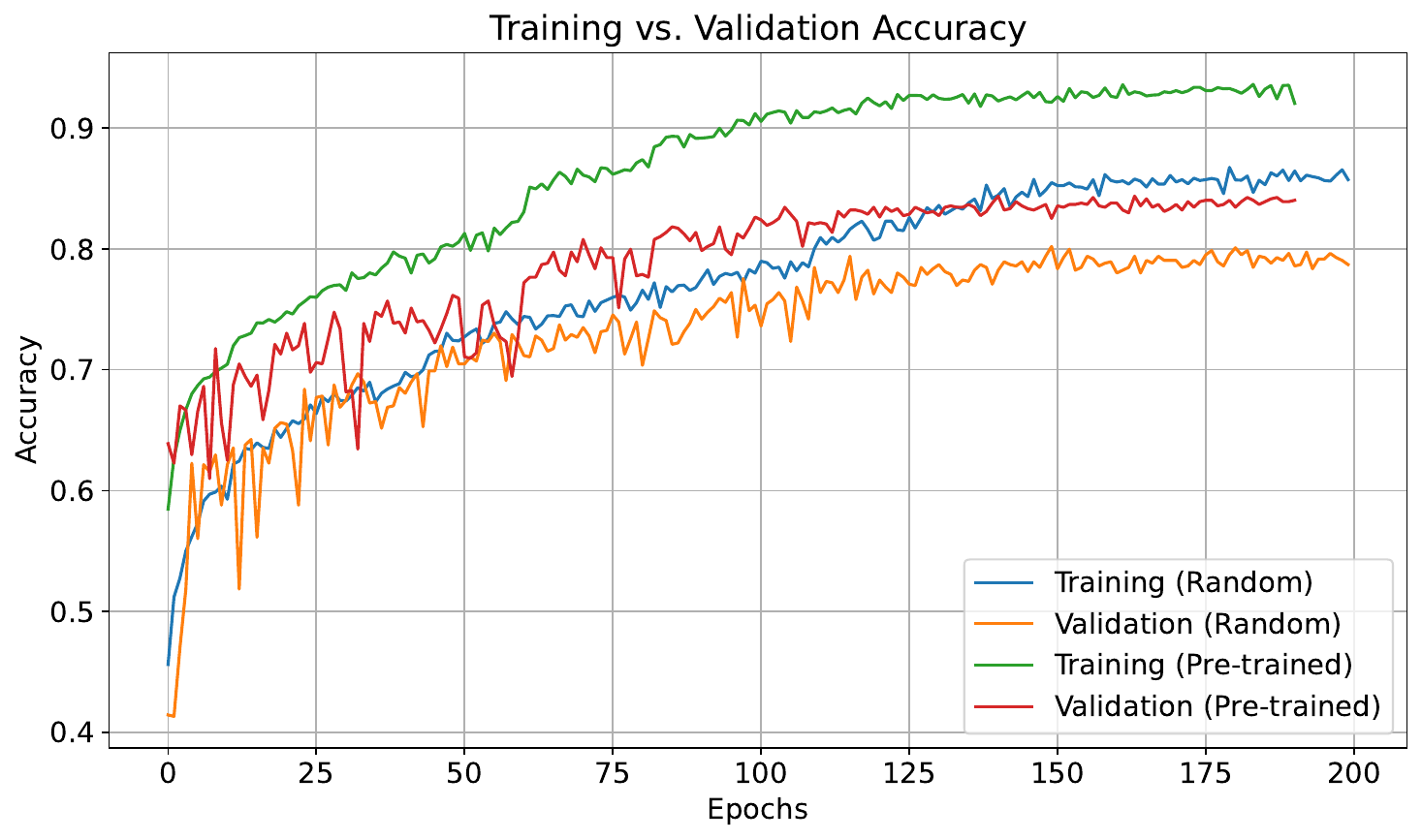}}
  \caption{Effects of random and pre-trained weight initialization on the train vs. validation accuracy over epochs}
  \label{fig:random_vs_pretrain_convergence} 
\end{figure*}

Table \ref{tab:performance-table} portrays the overall performance summary of our proposed model. The model demonstrated high classification accuracy across different datasets, with particularly strong performance on the PlantVillage and Coconut datasets, achieving accuracy rates of 99.57\% and 99.20\%, respectively. Among the CCMT datasets, the model exhibited robust results for CCMT-Cashew (95.04\%), CCMT-Cassava (94.34\%), and 92.76\% test accuracy for Sugarcane. These results suggest that the model generalizes well to different crop types, even when dealing with real-world datasets that may contain variations in lighting, background, and disease severity. The f1-scores for these datasets remain consistently high, with both macro and weighted averages exceeding 0.94 for CCMT-Cashew and CCMT-Cassava, further reinforcing the model’s balanced performance across classes. The performance on the CCMT-Maize and CCMT-Tomato datasets was comparatively lower, with accuracy rates of 81.05\% and 80.05\%, respectively. However, the macro and weighted precision, recall, and f1-scores for these datasets still indicate reasonably effective classification performance, suggesting that the model can be further improved with additional training data or enhanced preprocessing techniques. To grab the whole picture, the One-vs-rest (OvR) AUC metric provides crucial insight into the discriminatory power of our model. We observe that it produced above 0.99 score for all the data except Maize and Tomato. Overall, these findings indicate that the proposed MobilePlantViT achieves state-of-the-art performance on standardized datasets while maintaining strong generalizability across diverse agricultural datasets. 

To evaluate the impact of domain-specific pre-trained weight initialization on small-scale datasets, we initialized our model with weights pre-trained on the PlantVillage dataset and retrained it on the CCMT dataset. Our findings (presented in Table \ref{tab:pretrained-performance-table}) indicate that pre-training on relevant large-scale datasets (in our case PlantVillage) significantly enhances performance across small-scale datasets. The highest accuracy improvement was observed for CCMT-Maize and CCMT-Tomato, with increases of 4.00\% and 4.50\%, respectively. Precision, recall, and f1-score also followed a similar trend. These increases in the performance scores endorse that domain-specific pre-training provides valuable feature representations that generalize well to unseen datasets. Therefore, incorporating this strategy can positively impact the model performance, particularly for lightweight architecture and datasets with limited training samples. Figure \ref{fig:random_vs_pretrain_convergence} illustrates the training and validation accuracy and loss trends over epochs, highlighting the impact of random vs. pre-trained weight initialization. Models initialized with pre-trained weights show an immediate boost in both training and validation accuracy, leading to faster convergence in contrast to randomly initialized models. However, pre-trained weights can sometimes lead to overfitting (Figure \ref{1d}) and sufficient measures should be taken like regularization, early stopping, etc. Overall, leveraging pre-trained weights from relevant large-scale data enhances model performance and accelerates convergence which might come in handy for underperforming small datasets.

\subsection{Ablation Study}\label{subsec:ablation}
We conducted an ablation study on the proposed model to evaluate the contribution of its key components to classification performance. The experiments were performed on the CCMT-Tomato dataset, as it proved to be the most challenging dataset in our study. Table \ref{tab:ablation} summarizes the results of the ablation study. The findings reveal a substantial performance decline when either CBAM is removed or the GroupConv block formation is altered. In {Case 1}, where both CBAM and the proposed 1-2-4 GroupConv configuration were excluded, the model experienced the most significant drop in performance. Accuracy decreased by 9.32\%, macro precision by 6.07\%, weighted precision by 9.41\%, macro recall by 15.3\%, and macro f1-score by 13.47\% compared to the proposed model, as shown in Table \ref{tab:performance-table}. This demonstrates that both CBAM and the 1-2-4 GroupConv formation are crucial for enhancing feature extraction and representation learning. In {Case 2}, where CBAM was removed but the 1-2-4 GroupConv formation was retained, the performance drop was less severe than in Case 1 but still significant. Accuracy declined by 4.82\%, with macro precision, weighted precision, macro recall, and macro f1-score decreasing by 3.17\%, 4.85\%, 8.66\%, and 9.87\%, respectively. This suggests that while the 1-2-4 GroupConv structure plays a key role in feature extraction, CBAM further refines feature selection and improves classification performance. Lastly, in {Case 3}, where CBAM was retained but the GroupConv formation was changed to 1-1-1, the model experienced a moderate decline in performance. Accuracy decreased by 3.04\%, macro precision by 2.38\%, weighted precision by 2.88\%, macro-recall by 3.71\%, and macro f1-score by 3.37\%. These results highlight that while CBAM enhances feature representation, the hierarchical structure of the 1-2-4 GroupConv formation is a significant contributor to robust classification performance. Overall, the ablation study confirms that both the CBAM attention mechanism and the hierarchical 1-2-4 GroupConv block formation contribute synergistically to the model’s performance. Removing either component results in performance degradation, with the worst-case scenario occurring when both CBAM and the 1-2-4 configuration are absent. Thus, our findings support the inclusion of both components in the proposed model to achieve optimal performance in plant disease classification.

\begin{table*}[htpb]
    \centering
    \caption{Summary of the ablation study highlighting the key contributions of the core components of the model}
    \label{tab:ablation}
    \begin{tabular}{ccccc|cc|cc}
        \hline\hline
        \multirow{2}{*}{\textbf{CBAM}} & \multirow{2}{*}{\textbf{GroupConv}} 
        & \multirow{2}{*}{\textbf{Accuracy (\%)}} 
        & \multicolumn{2}{c}{\textbf{Precision (\%)}} & \multicolumn{2}{c}{\textbf{Recall (\%)}} & \multicolumn{2}{c}{\textbf{F1-score}} \\ 
        \cline{4-9}
        & & & \textbf{Macro} & \textbf{Weighted} & \textbf{Macro} & \textbf{Weighted} & \textbf{Macro} & \textbf{Weighted} \\ 
        \hline
        {\xmark} & 1-1-1 & 72.59 (9.32$\downarrow$) & 74.75 (6.07$\downarrow$) & 72.35 (9.41$\downarrow$) & 64.68 (15.3$\downarrow$) & 72.59 (9.32$\downarrow$) & 0.6708 (13.47\%$\downarrow$) & 0.7079 (11.22\%$\downarrow$) \\
        {\xmark} & 1-2-4 & 75.23 (4.82$\downarrow$) & 76.41 (3.17$\downarrow$) & 75.01 (4.85$\downarrow$) & 67.70 (8.66$\downarrow$) & 75.23 (4.82$\downarrow$) & 0.6989 (9.87\%$\downarrow$) & 0.7385 (7.38\%$\downarrow$) \\
        {\checkmark} & 1-1-1 & 77.64 (3.04$\downarrow$) & 77.69 (2.38$\downarrow$) & 77.56 (2.88$\downarrow$) & 73.53 (3.71$\downarrow$) & 77.64 (3.04$\downarrow$) & 0.7493 (3.37\%$\downarrow$) & 0.7721 (3.16\%$\downarrow$) \\
        \textbf{\checkmark} & \textbf{1-2-4} & \textbf{80.05} & \textbf{79.58} & \textbf{79.86} & \textbf{76.36} & \textbf{80.05} & \textbf{0.7754} & \textbf{0.7973}\\
        \hline\hline
    \end{tabular}
\end{table*}

\subsection{Error Analysis}\label{subsec:error}
\begin{figure} 
    \centering
  \subfloat[CCMT-Maize Confusion Matrix\label{2a}]{%
        \includegraphics[width=0.49\linewidth]{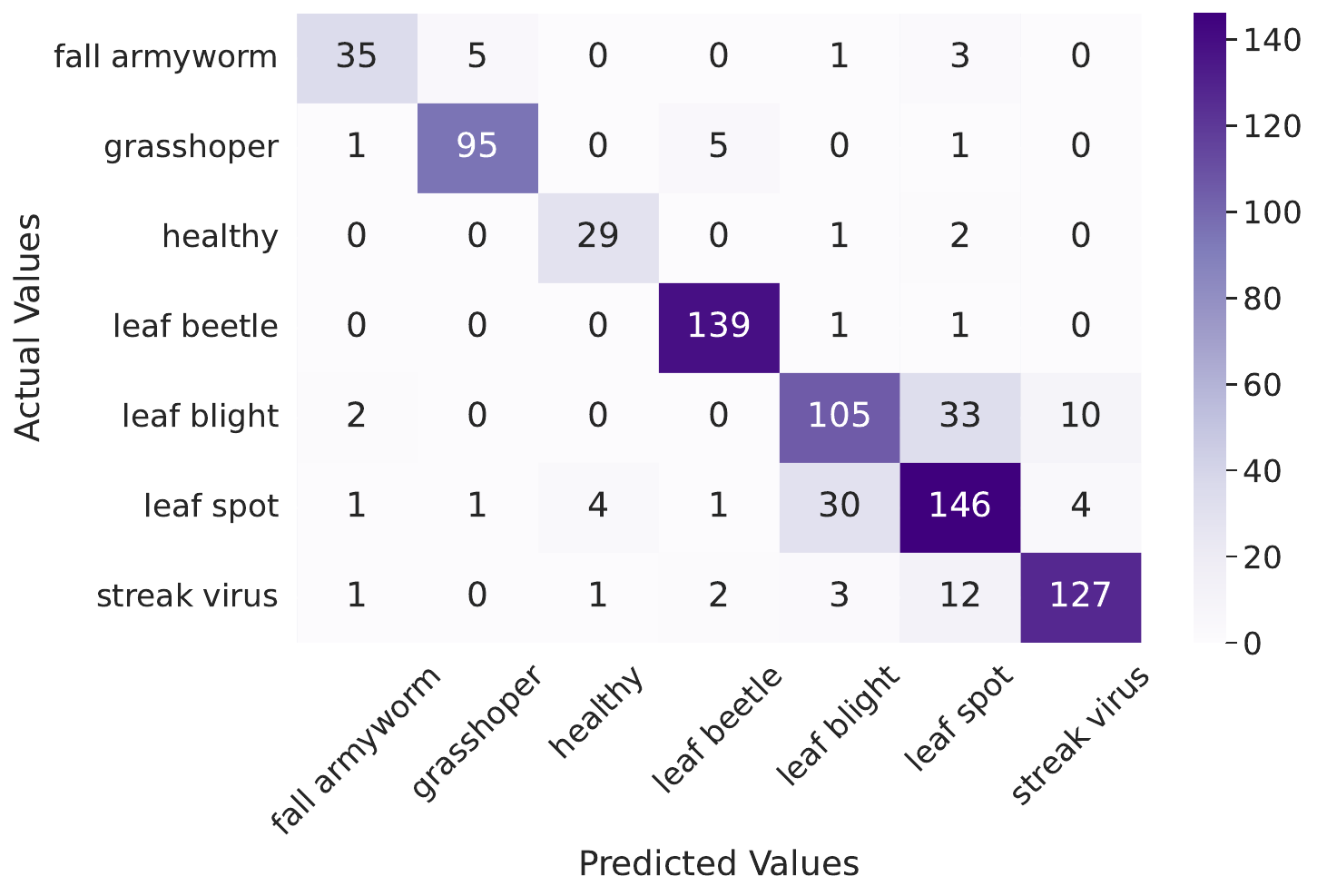}}
    \hfill
  \subfloat[CCMT-Tomato Confusion Matrix\label{2b}]{%
        \includegraphics[width=0.49\linewidth]{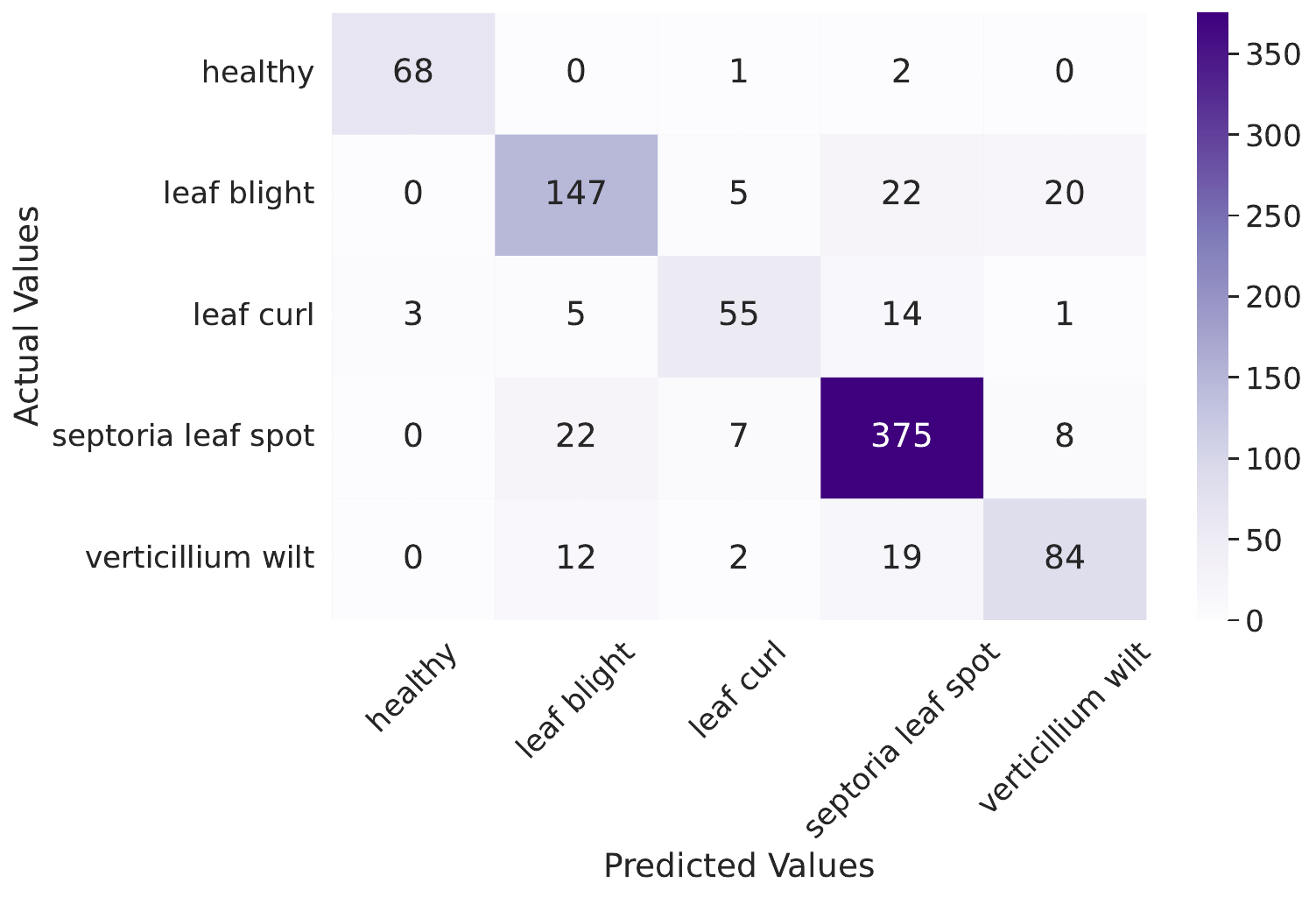}}
    \\
  \subfloat[CCMT-Maize Misclassified Samples\label{2c}]{%
        \includegraphics[width=0.99\linewidth]{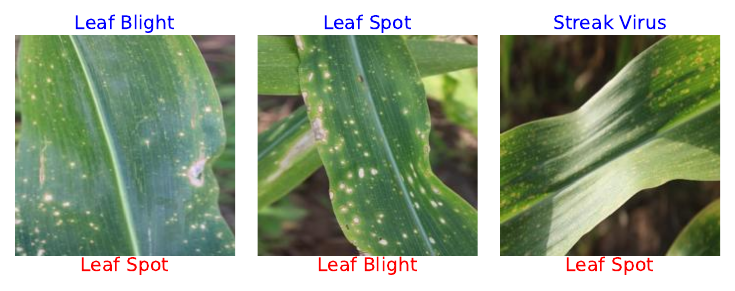}}
    \\    
  \subfloat[CCMT-Tomato Misclassified Samples\label{2d}]{%
        \includegraphics[width=0.99\linewidth]{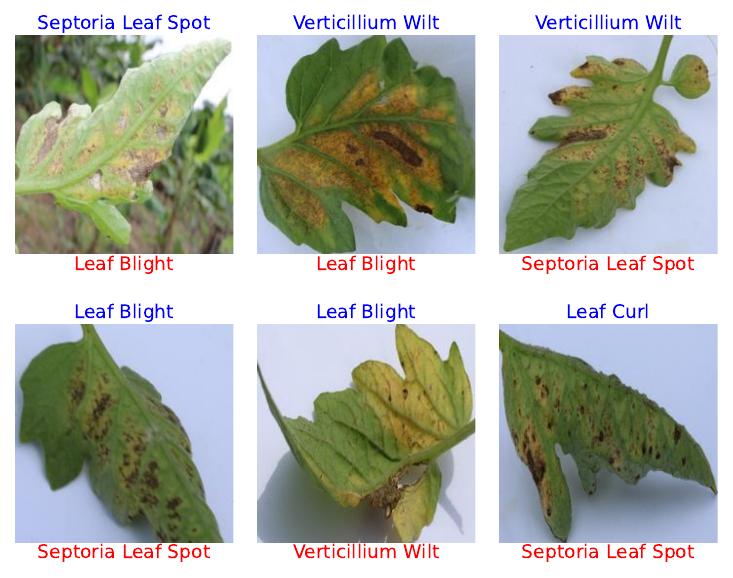}}    
  \caption{Confusion matrices representing the true positives (TP), false positives (FP), true negatives (TN), and false negatives (FN) for Maize and Tomato, along with misclassified samples}
  \label{fig:error-analysis} 
\end{figure}

The most frequent misclassification for Maize occurred between \textit{leaf blight, leaf spot}, and \textit{streak virus}, and Tomatos involved \textit{septorial leaf spot, verticillium wilt, leaf blight}, and \textit{leaf curl}, observed via confusion matrix analysis presented in Figure \ref{2a}, \ref{2b}. As depicted in Figure \ref{2c}, \ref{2d}, one key reason for these misclassifications is the high similarity in the visual patterns of the affected areas. While lightweight models typically involve a trade-off between accuracy and resource efficiency, it is noteworthy that our model achieved competitive accuracy on the PlantVillage dataset (see Table \ref{tab:performance-table}), which includes a subset of Tomato disease instances. The smaller training size and data quality likely played a significant role in these comparatively lower performances. To improve further, utilizing extensive augmentations to introduce more diverse large-scale training data, applying targeted augmentations on hard-to-classify samples, and domain-specific pre-training on large datasets could be beneficial. Additionally, contrastive learning and hierarchical classification approaches may also be useful in applicable contexts.

\subsection{Performance Comparison With Equivalent ViTs}\label{subsec:perform-comp}

\begin{table*}[htpb]
    \centering
    \caption{Performance comparison of lightweight ViTs having equivalent parameters}
    \label{tab:sota-table}
    \begin{tabular}{lclccc|cc|cc}
        \hline\hline
        \multirow{2}{*}{\textbf{Model}} 
        & \multirow{2}{*}{\textbf{Params (M)}}  & \multirow{2}{*}{\textbf{Data}} & \textbf{Accuracy (\%)} 
        & \multicolumn{2}{c}{\textbf{Precision (\%)}} & \multicolumn{2}{c}{\textbf{Recall (\%)}} & \multicolumn{2}{c}{\textbf{F1-score}} \\ 
        \cline{5-10}
        & & & & \textbf{Macro} & \textbf{Weighted} & \textbf{Macro} & \textbf{Weighted} & \textbf{Macro} & \textbf{Weighted} \\ 
        \hline
        \multirow{4}{*}{MobileViTv1-XXS} & \multirow{4}{*}{0.95} & CCMT-Cashew & 94.02 & 94.82 & 94.01 & 94.88 & 94.02 & 0.9484 & 0.9401 \\
                                         &  & CCMT-Cassava & 91.25 & 91.18 & 91.32 & 91.90 & 91.25 & 0.9152 & 0.9126 \\
                                         &  & CCMT-Maize & 74.56 & 74.80 & 75.49 & 72.67 & 74.56 & 0.7324 & 0.7475 \\
                                         &  & CCMT-Tomato & 69.95 & 70.76 & 69.58 & 65.06 & 69.95 & 0.6685 & 0.6921 \\\cline{1-10}
        \multirow{4}{*}{MobileViTv2-050} & \multirow{4}{*}{1.12} & CCMT-Cashew & 94.43 & 95.22 & 94.43 & 94.76 & 94.43 & 0.9497 & 0.9441 \\
                                         &  & CCMT-Cassava & 93.81 & 94.37 & 93.82 & 93.93 & 93.81 & 0.9414 & 0.9381 \\
                                         &  & CCMT-Maize & 77.43 & 78.17 & 77.89 & 77.66 & 77.43 & 0.7783 & 0.7761 \\
                                         &  & CCMT-Tomato & 73.28 & 73.83 & 72.75 & 68.35 & 73.28 & 0.7018 & 0.7246 \\\cline{1-10}
        \multirow{4}{*}{MobilePlantViT}  & \multirow{4}{*}{0.69} & CCMT-Cashew & \textbf{95.04} & \textbf{95.71} & \textbf{95.02} & \textbf{95.78} & \textbf{95.04} & \textbf{0.9574} & \textbf{0.9502} \\
                                         &  & CCMT-Cassava & \textbf{94.34} & \textbf{94.46} & \textbf{94.40} & \textbf{94.41} & \textbf{94.36} & \textbf{0.9442} & \textbf{0.9436} \\  
                                         &  & CCMT-Maize & \textbf{81.05} & \textbf{82.87} & \textbf{81.41} & \textbf{81.42} & \textbf{81.05} & \textbf{0.8185} & \textbf{0.8110} \\
                                         &  & CCMT-Tomato & \textbf{80.05} & \textbf{79.58} & \textbf{79.86} & \textbf{76.36} & \textbf{80.05} & \textbf{0.7754} & \textbf{0.7973} \\ 
                                         
        \hline\hline
    \end{tabular}
\end{table*}

We conducted a comparative performance analysis against the lightweight versions of the MobileViT models with equivalent parameter constraints. We picked the MobileViTv1-XXS and the MobileViTv2-050 versions. The results are reported in Table \ref{tab:sota-table} provides a summarized assessment. Our proposed model, MobilePlantViT, consistently outperformed MobileViT-XXS and MobileViTV2-050 across the experimental datasets. Despite having fewer parameters (0.69M) than both MobileViT-XXS (0.95M) and MobileViTV2-050 (1.12M), MobilePlantViT achieved the highest accuracy, precision, recall, and f1-scores. Notably, the performance improvement was more pronounced for the CCMT-Maize and CCMT-Tomato datasets, the challenging data segments of our experiments. Our model achieved a peak accuracy of 80.05\%-83.60\% on the CCMT-Tomato data, outperforming the MobileViTV2-050 producing an accuracy of 73.28\% and MobileViT-XXS by 69.95\%. For CCMT-Maize, our model achieved 81.05\% accuracy with random weight initialization and 84.29\% with pre-trained weight initialization, significantly outperforming MobileViTV2-050 (77.43\%) and MobileViT-XXS (74.56\%). Similar trends of producing lower performance scores were followed in the CCMT-Cashew and CCMT-Cassava as well. This significant performance superiority of our proposed model is reflected in all performance metrics, including macro and weighted precision, recall, and f1-score. 


\section{Conclusion and Future Direction}\label{sec:conclusion}
In this study, we introduce MobilePlantViT, a novel DL model for effective and generalized plant disease image classification. Our model is computationally lightweight, making it well-suited for mobile and resource-constrained edge devices, with the broader goal of enhancing the accessibility and scalability of smart agricultural systems. To achieve efficiency without compromising accuracy, MobilePlantViT first extracts essential features via a stack of group convolutions that are fused with convolutional attention modules, progressively downsampling feature representations to reduce computational complexity before passing them to the encoder. Unlike conventional multi-head self-attention, which operates with quadratic complexity, our encoder leverages a self-attention mechanism with linear complexity, significantly improving scalability. Extensive experiments across multiple plant disease datasets, accompanied by an in-depth performance analysis from various perspectives demonstrate that MobilePlantViT achieves not only promising classification performance but also generalizes effectively across different instances of plant disease. In the future, further research could focus on adapting MobilePlantViT to other agricultural image classification tasks, expanding its applicability within precision agriculture. Additionally, exploring domain-specific pre-training could lead to improved performance and better generalizability, particularly for handling rare diseases or diverse environmental conditions, ultimately enhancing the model's robustness across various agricultural domains.


\bibliographystyle{IEEEtran}
\bibliography{ref}



\end{document}